\pdfoutput=1
\documentclass[pdflatex,sn-basic,Numbered]{sn-jnl}

\usepackage{graphicx}
\graphicspath{{figures/}}
\usepackage{amsmath,amssymb}
\usepackage{booktabs}
\usepackage{multirow}
\usepackage{tabularx}
\usepackage{array}
\usepackage{makecell}
\usepackage{subcaption}
\usepackage{microtype}
\usepackage{url}
\usepackage{placeins}
\usepackage{float}
\usepackage{enumitem}
\usepackage{xcolor}
\usepackage{tikz}
\usetikzlibrary{arrows.meta,positioning,fit,calc}
\usepackage[ruled,vlined,linesnumbered]{algorithm2e}

\newcolumntype{Y}{>{\centering\arraybackslash}X}
\newcolumntype{L}{>{\raggedright\arraybackslash}X}
\newcommand{\best}[1]{\textbf{#1}}
\newcommand{\pmv}[2]{#1$\pm$#2}

\begin{document}

\title[DRIL for image-based continuous control]{Disagreement-Regularized Imitation Learning for Image-Based Continuous Control with Gaussian and Beta Policies}

\author[1]{\fnm{Irving Giovani Bronzatti} \sur{Petrazzini}}
\email{irving.petrazzini@posgrad.ufsc.br}

\author*[1]{\fnm{Eric Aislan} \sur{Antonelo}}
\email{eric.antonelo@ufsc.br}

\affil[1]{\orgdiv{Department of Automation and Systems Engineering}, \orgname{Federal University of Santa Catarina}, \orgaddress{\city{Florian\'opolis}, \state{Santa Catarina}, \country{Brazil}}}

\abstract{\textbf{Purpose:} Behavior cloning can accumulate errors when a learned controller visits states outside the demonstrated distribution. This study evaluates whether Disagreement-Regularized Imitation Learning (DRIL), which converts disagreement among cloned policies into a reinforcement-learning reward, improves image-based continuous control. \textbf{Methods:} A controlled CarRacing study combines Gaussian and Beta learner policies, demonstrations from either a clipped Gaussian expert or an intrinsically bounded Beta expert, one or 20 trajectories, deterministic and stochastic evaluation, and three retained stages: behavior cloning, the highest 10-episode training-score checkpoint, and the final DRIL checkpoint. The disagreement ensemble contains five Gaussian policies in every variant. Each retained policy is evaluated over 100 procedurally generated episodes. \textbf{Results:} Score-selected DRIL produced its largest gains in the few-demonstration setting, improving over the strongest behavior-cloning mean by 61\% with clipped-action demonstrations and by 112\% with bounded-action demonstrations. With 20 trajectories, the advantage of DRIL narrowed; in the bounded-action regime, Beta behavior cloning remained about 7\% above the best DRIL checkpoint. The experiments also show that the informativeness of the disagreement reward changes with the learner representation and training stage. \textbf{Conclusion:} DRIL can substantially improve few-demonstration visual continuous control, while bounded Beta policies provide strong behavior-cloning performance when more demonstrations are available. The results highlight the joint importance of learner support, ensemble response, and checkpoint selection.}


\maketitle

\section{Introduction}\label{sec:intro}

Learning control directly from demonstrations is a long-standing approach to autonomous driving. Early work mapped camera observations to steering commands with a neural network \cite{pomerleau1989}, and later end-to-end systems scaled this idea with deep visual encoders and larger datasets \cite{bojarski2016,codevilla2018}. The simplest form, behavior cloning (BC), treats expert state--action pairs as supervised examples. BC is inexpensive compared with online reinforcement learning (RL), does not require a manually designed task reward, and can reuse logged demonstrations. Its principal weakness is distribution shift: training minimizes error under the expert's state distribution, while deployment generates states under the learned policy. Small errors therefore compound and can lead to states that were never demonstrated \cite{ross2010,ross2011}.

Interactive imitation-learning methods reduce this mismatch by obtaining supervision on states visited by the learner. DAgger aggregates labels from an expert during policy rollouts \cite{ross2011}; active-learning variants query a teacher when the policy is uncertain \cite{hussein2018}; and hybrid RL--IL methods use environment interaction to refine a cloned policy \cite{kumar2024,lu2022}. These approaches can improve recovery, but they may require repeated expert access, a task reward, or a learned reward model.

Disagreement-Regularized Imitation Learning (DRIL) offers a different mechanism \cite{brantley2020}. An ensemble of policies is trained by BC on bootstrapped expert data. Agreement among the ensemble members is expected on demonstrated states, whereas disagreement is treated as evidence that a rollout has left the demonstrated distribution. The ensemble variance is thresholded to form a binary reward for PPO, and BC updates are interleaved with policy-gradient updates. DRIL therefore replaces explicit task-reward design with a reward induced by imitation uncertainty.

The original DRIL evaluation considered image observations with discrete actions (Atari) and low-dimensional observations with continuous actions. End-to-end autonomous control combines both challenges: high-dimensional visual input and continuous bounded actions. This combination raises two questions that are not independent. First, does ensemble disagreement remain useful when the learner's action is sampled from a continuous distribution? Second, how does the support of that distribution affect both BC and the disagreement signal?

Gaussian policies are common in continuous-control RL, but their infinite support requires clipping when actuators are bounded. The clipping operation changes the executed action and can bias policy-gradient estimation \cite{chou2017}. A Beta policy has finite support and generates valid actions by construction. Previous work showed that replacing the Gaussian distribution with the Beta distribution improves PPO in CarRacing \cite{petrazzini2021}. In imitation learning, however, a stronger bounded policy can change the state visitation distribution and the scale of ensemble variance. It can therefore improve BC while simultaneously weakening the reward on which DRIL depends.

The objective of this study is to determine whether disagreement regularization can improve behavior cloning in image-based continuous control, and to analyze how learner-policy support, expert-action bounds, demonstration count, evaluation mode, and checkpoint selection affect the result. The work advances the original DRIL evaluation by studying the previously untested combination of high-dimensional visual observations and continuous bounded actions, while systematically analyzing the interaction among policy representation, demonstration source, dataset size, and checkpoint selection.

The main contributions are:

\begin{enumerate}[leftmargin=*,itemsep=2pt]
    \item an adaptation and evaluation of DRIL for stacked image observations and two-dimensional continuous bounded actions;
    \item a factorial comparison of Gaussian and Beta learner policies using demonstrations generated by either a clipped Gaussian expert or an intrinsically bounded Beta expert, with one and 20 trajectories;
    \item a systematic comparison of BC, score-selected transient DRIL checkpoints, and final DRIL checkpoints, exposing the dependence of the reported performance on model-selection protocol;
    \item an empirical analysis of training-stage dynamics and of the disagreement-reward response when a Beta learner is paired with the fixed Gaussian ensemble;
    \item complete deterministic and stochastic 100-episode evaluations for every retained policy configuration.
\end{enumerate}

The strongest result is not that one policy distribution dominates universally. Score-selected DRIL obtains the highest stochastic mean in three of four demonstration regimes, whereas Beta BC is strongest when 20 bounded-action trajectories are available as a training set. The evidence instead shows that the usefulness of disagreement regularization depends on the policy representation, the informativeness of the thresholded uncertainty reward, and the checkpoint that is retained.

\section{Related work}\label{sec:related}

\subsection{Imitation learning and covariate shift}

BC learns a direct mapping from observation to action and is widely used because it reduces imitation to standard supervised learning. Its closed-loop weakness motivated reductions from imitation learning to online learning and dataset aggregation \cite{ross2010,ross2011}.
More recent theoretical analysis has revisited the commonly assumed
performance gap between offline behavior cloning and online imitation
learning, showing that the horizon dependence of BC can be substantially
improved under appropriate assumptions \cite{foster2024}.
Causal confusion is a related problem: a policy can exploit correlations in the demonstrations that do not remain valid after intervention or deployment \cite{dehaan2019}. Adversarial imitation methods such as GAIL instead match expert occupancy measures through a discriminator, but introduce adversarial optimization and usually require substantial environment interaction \cite{ho2016}. BC-augmented GAIL methods attempt to improve sample efficiency and stability \cite{jena2021,Couto2024}.

Hussein et al. used visual demonstrations for 3D navigation and refined the cloned policy through active learning on uncertain states \cite{hussein2018}. Celemin and Kober explicitly separated epistemic uncertainty from demonstration ambiguity when deciding when corrective or evaluative human feedback was needed \cite{celemin2023}. Kumar combined imitation guidance and RL for navigation in dynamic environments \cite{kumar2024}. Chen and Xuan used adversarial inverse RL to infer a reward for visual continuous drone navigation \cite{chen2026}. These studies share the objective of moving beyond static cloning, but differ in the source of corrective information. DRIL neither requests a new expert label nor attempts to recover the expert's latent task reward; it constructs a surrogate reward from disagreement among cloned policies.

The distribution-shift problem also connects to offline RL, where a learned policy may select actions poorly represented in a fixed dataset. A recent survey organizes solutions around policy constraints, uncertainty penalties, and out-of-distribution action control \cite{riahi2026}. DRIL is not an offline-RL algorithm because it interacts with the environment during PPO training, but its binary disagreement reward acts as a constraint that encourages the policy to remain near the demonstrated distribution.

\begin{table}[t]
\caption{Qualitative comparison of methods that address imitation-learning distribution shift. ``Expert queries'' indicates whether additional labels are required after the initial demonstrations. ``Task reward'' indicates use of the environment reward during policy improvement.}
\label{tab:related_comparison}
\centering
\small
\begin{tabularx}{\textwidth}{@{}lYYL@{}}
\toprule
Method & Expert queries & Task reward & Corrective mechanism \\
\midrule
BC & No & No & None; supervised fitting on the fixed dataset \\
DAgger \cite{ross2011} & Yes & No & Aggregates expert labels on learner-visited states \\
GAIL \cite{ho2016} & No & No & Adversarial occupancy-measure matching \\
Hybrid RL--IL \cite{kumar2024,lu2022} & Varies & Yes & Refines imitation with environment rewards \\
Diffusion Meets DAgger \cite{zhang2024dmd} & No new labels & No & Synthesizes examples near failure states \\
DRIL \cite{brantley2020} & No & No & Thresholded disagreement among cloned policies \\
\bottomrule
\end{tabularx}
\end{table}

\subsection{Visual imitation learning for autonomous control}

Autonomous-driving imitation learning has evolved from direct camera-to-control mappings \cite{pomerleau1989,bojarski2016} to conditional policies and privileged teachers \cite{codevilla2018,zhang2021}.
Behavior cloning with Bird's-Eye View representations
has also been investigated under limited demonstration data in simulated urban driving
\cite{Antonelo2024BC}.
More recently, hybrid learning-based controllers have combined neural
control with model-based expertise and online imitation learning in CARLA,
providing another approach to adapting autonomous-driving policies from
expert demonstrations \cite{hiremath2026}.
Other work combines imitation and RL to improve robustness in difficult driving scenarios \cite{lu2022}. Closed-loop evaluation is essential because low supervised error does not guarantee route completion or recovery from deviations. Modern benchmarks such as Bench2Drive evaluate multiple closed-loop abilities and scenario types in CARLA \cite{dosovitskiy2017,jia2024}. CarRacing remains useful for controlled experiments because tracks are generated procedurally, policy execution is closed loop, observations are images, and actions are continuous.

\subsection{Uncertainty and ensemble disagreement}

Ensembles provide a practical estimate of epistemic uncertainty. In DRIL, multiple BC policies are trained from different initializations and bootstrapped data. Their prediction variance is expected to be small on demonstrated samples and larger elsewhere \cite{brantley2020}. A high quantile of uncertainty on the expert data defines a threshold; crossing it produces a negative reward. Empirical follow-up studies have shown that deep imitation methods, including DRIL, are sensitive to implementation choices and to the interaction between BC and RL updates \cite{sasaki2021,arulkumaran2021}.

Recent work on runtime monitoring of generative imitation policies further shows that learned policies can exhibit qualitatively different failure modes, motivating complementary indicators beyond a single uncertainty or consistency signal \cite{Agia2025}.

The behavior of ensemble variance depends on both the policy representation and the states visited during rollout. Celemin and Kober's distinction between epistemic and aleatoric uncertainty \cite{celemin2023} is therefore directly relevant. The experiments below analyze this interaction explicitly: for Beta learners, the fixed Gaussian ensemble often remains below its expert-data variance threshold, producing an almost uniformly positive reward and revealing how the reward signal changes across learner representations.

\subsection{Continuous policy representations}

A Gaussian policy is convenient for policy gradients but has unbounded support. Clipping a sampled action to a bounded interval changes the action applied by the environment and can introduce gradient bias \cite{chou2017}. The Beta distribution has finite support and avoids invalid samples. Petrazzini and Antonelo previously compared Gaussian and Beta PPO policies in LunarLanderContinuous and CarRacing, finding better stability and performance for the Beta parameterization \cite{petrazzini2021}. That prior PPO contribution is used here only to provide a bounded expert and motivate the learner parameterization; it is not claimed as a new contribution of this paper.

Recent imitation-learning policies are considerably more expressive than a unimodal Gaussian or Beta actor. Implicit Behavioral Cloning represents the policy with an energy function \cite{florence2022};
recent work has also explored energy-based implicit
behavior cloning for multimodal vehicle navigation in simulated cities \cite{Antonelo2025IBC};
Behavior Transformers model multimodal action distributions through discrete action clusters and residuals \cite{shafiullah2022}; Diffusion Policy generates action sequences through conditional denoising \cite{chi2023}; and VQ-BeT uses latent action tokens for multimodal behavior generation \cite{lee2024}. Diffusion Meets DAgger addresses covariate shift by synthesizing examples near failure states rather than querying a teacher \cite{zhang2024dmd}. These methods provide a modern context for the explicit Gaussian and Beta policy representations evaluated in this study.

\section{Method}\label{sec:method}

\subsection{Problem formulation and behavior cloning}

Let $\mathcal{D}=\{(s_i,a_i)\}_{i=1}^{N}$ be a set of expert state--action pairs. A policy $\pi_\theta(a\mid s)$ is trained to reproduce the expert. For continuous actions, BC minimizes the mean-squared error between the expert action and the policy mean,
\begin{equation}
\mathcal{L}_{\mathrm{BC}}(\theta)=\frac{1}{N}\sum_{i=1}^{N}\left\|\mu_\theta(s_i)-a_i\right\|_2^2.
\label{eq:bc}
\end{equation}
The data are split randomly into 80\% training and 20\% validation samples. Training stops after 20 epochs without improvement in validation loss, up to a maximum of 2000 epochs, and the best validation checkpoint initializes DRIL.

For CarRacing, the action is two dimensional. The first component is steering in $[-1,1]$. Throttle and brake are merged into a second component in $[-1,1]$, where $-1$ is full brake, $+1$ is full throttle, and zero applies neither. The merged action avoids simultaneous acceleration and braking and matches the policy used to generate the expert demonstrations.

\subsection{Disagreement ensemble}

An ensemble $\Pi_E=\{\pi_{\phi_j}\}_{j=1}^{E}$ with $E=5$ is trained by BC. Each member starts from a different random initialization and receives a bootstrap sample drawn with replacement, with the same size as the original demonstration dataset. Ensemble members use the Gaussian policy architecture and minimize mean-squared error between their predicted action means and the expert actions. They are trained for exactly 2000 epochs without early stopping so that they fit the training set while retaining different extrapolations away from it. This Gaussian ensemble is held fixed in design for all main experiments: in the DRIL-Beta variants, only the learner actor is changed to a Beta policy.

For a continuous action, each ensemble member predicts a mean vector
$m_j(s)\in\mathbb{R}^{d}$. The empirical mean and covariance are
\begin{align}
\bar m(s) &= \frac{1}{E}\sum_{j=1}^{E}m_j(s),\\
\Sigma_E(s) &=
\frac{1}{E}\sum_{j=1}^{E}
\big(m_j(s)-\bar m(s)\big)
\big(m_j(s)-\bar m(s)\big)^\top.
\end{align}

Following the continuous-action implementation used in this work,
the covariance matrix is mapped to an action-dependent scalar through the
quadratic form
\begin{equation}
u(s,a)=a^\top\Sigma_E(s)a.
\label{eq:uncertainty}
\end{equation}

Equation~\eqref{eq:uncertainty} is the variance obtained after projecting
the ensemble predictions onto the axis defined by $a$, because
$\operatorname{Var}(a^\top M)=a^\top\Sigma_E(s)a$.

For the two-dimensional action
$a=[a_{\mathrm{steer}},a_{\mathrm{long}}]^\top$, this becomes
\begin{equation}
u(s,a)=
a_{\mathrm{steer}}^2\Sigma_{11}
+2a_{\mathrm{steer}}a_{\mathrm{long}}\Sigma_{12}
+a_{\mathrm{long}}^2\Sigma_{22}.
\end{equation}

Thus, only disagreement aligned with $a$ contributes to the scalar
uncertainty measure, including the cross-covariance between steering and
longitudinal control. Since the implementation uses the raw action vector
rather than a unit vector, the score depends on both the action direction
and its magnitude.

The 98th percentile of $u(s,a)$ over expert training samples defines
$q_{0.98}$. The clipped uncertainty cost and equivalent reward are
\begin{align}
C_U^{\mathrm{clip}}(s,a)&=
\begin{cases}
-1,&u(s,a)\le q_{0.98},\\
+1,&u(s,a)>q_{0.98},
\end{cases}\\
r_U(s,a)&=-C_U^{\mathrm{clip}}(s,a).
\label{eq:reward}
\end{align}
Thus, ensemble agreement yields $+1$ and disagreement yields $-1$, as reward signals.

\subsection{DRIL optimization}

\begin{figure}[!b]
\centering
\includegraphics[width=0.92\textwidth]{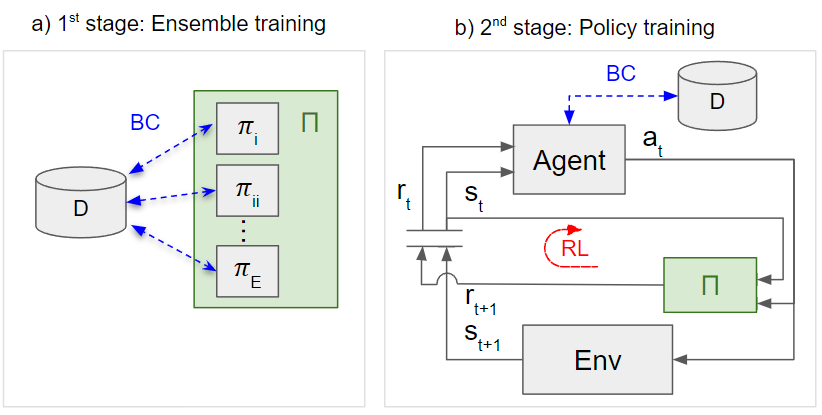}
\caption{Two-stage DRIL workflow used in this work. In stage (a), the expert demonstration dataset $\mathcal{D}$ trains the ensemble $\Pi$ by behavior cloning; in the present experiments, $\Pi$ contains five Gaussian policies trained on bootstrap samples. In stage (b), a BC-initialized learner interacts with the environment. Ensemble disagreement provides the binary reward for reinforcement-learning updates, while behavior-cloning updates continue to use the original demonstrations.}
\label{fig:dril_overview}
\end{figure}

The BC-initialized learner interacts with the environment using the uncertainty reward in Eq.~\eqref{eq:reward}. PPO optimizes the clipped surrogate objective \cite{schulman2017}, with advantages estimated by generalized advantage estimation \cite{schulman2015gae}. One BC update is followed by one PPO policy-gradient update for each optimization minibatch, preserving direct imitation while encouraging the rollout distribution to remain in low-disagreement regions. Figure~\ref{fig:dril_overview} summarizes the procedure.

\begin{algorithm}[H]
\caption{DRIL training and checkpoint reporting}
\label{alg:dril}
\KwIn{Expert data $\mathcal{D}$; ensemble size $E=5$; uncertainty quantile $0.98$}
Train learner $\pi$ by BC with an 80/20 split and retain the best validation checkpoint $\pi_{\mathrm{BC}}$\;
Set $\pi\leftarrow\pi_{\mathrm{BC}}$\;
\For{$e\leftarrow1$ \KwTo $E$}{
  Draw $\mathcal{D}_e$ from $\mathcal{D}$ with replacement, with $|\mathcal{D}_e|=|\mathcal{D}|$\;
  Train Gaussian ensemble member $\pi_e$ on $\mathcal{D}_e$ for 2000 epochs\;
}
Compute $q_{0.98}$ from the ensemble uncertainty on expert training samples\;
Initialize score-selected checkpoint $\pi_{\mathrm{peak}}\leftarrow\pi_{\mathrm{BC}}$ and best 10-episode mean $M\leftarrow-\infty$\;
\While{the DRIL interaction budget is not exhausted}{
  Collect a 2048-step rollout with $\pi$ using the binary uncertainty reward\;
  \ForEach{optimization minibatch}{
    Perform one gradient update to minimize $J_{\mathrm{BC}}(\pi)$ using a minibatch from $\mathcal{D}$\;
    Perform one PPO policy-gradient update using the disagreement rewards from the rollout\;
  }
  \If{the latest 10-episode mean task score is at least $M$}{
    $\pi_{\mathrm{peak}}\leftarrow\pi$; update $M$\;
  }
}
Set $\pi_{\mathrm{final}}\leftarrow\pi$\;
\KwOut{Behavior-cloning policy $\pi_{\mathrm{BC}}$; task-score-selected DRIL checkpoint $\pi_{\mathrm{peak}}$; final DRIL policy $\pi_{\mathrm{final}}$}
\end{algorithm}

The algorithm returns policies from three stages of the same learner. The behavior-cloning checkpoint $\pi_{\mathrm{BC}}$ is retained before any disagreement-regularized policy-gradient update and serves as the imitation baseline. During DRIL training, $\pi_{\mathrm{peak}}$ is replaced whenever the moving mean of the latest 10 CarRacing task scores exceeds its previous maximum; it therefore represents the strongest transient checkpoint according to the environment score rather than the ensemble reward. The final policy $\pi_{\mathrm{final}}$ is the learner obtained after the complete DRIL interaction budget, irrespective of whether an earlier checkpoint achieved a higher score. Reporting all three policies separates the initial cloned solution, the best task-score-selected intermediate policy, and the policy produced at the prescribed end of training.

The key distinction in this paper is between two retained DRIL checkpoints:
\begin{itemize}[leftmargin=*]
    \item \textbf{DRIL-final}: the policy after approximately 3 million environment steps (about 3000 episodes);
    \item \textbf{DRIL-peak}: the policy stored whenever the moving average of the previous 10 environment task scores exceeds its earlier maximum.
\end{itemize}
DRIL-peak uses the true CarRacing task score for checkpoint selection and is reported alongside both the BC initialization and the final DRIL checkpoint so that the complete training trajectory can be compared.

\subsection{Gaussian and Beta learner policies}

\begin{figure}[t]
\centering
\begin{subfigure}[t]{0.96\textwidth}
\centering
\includegraphics[width=\linewidth]{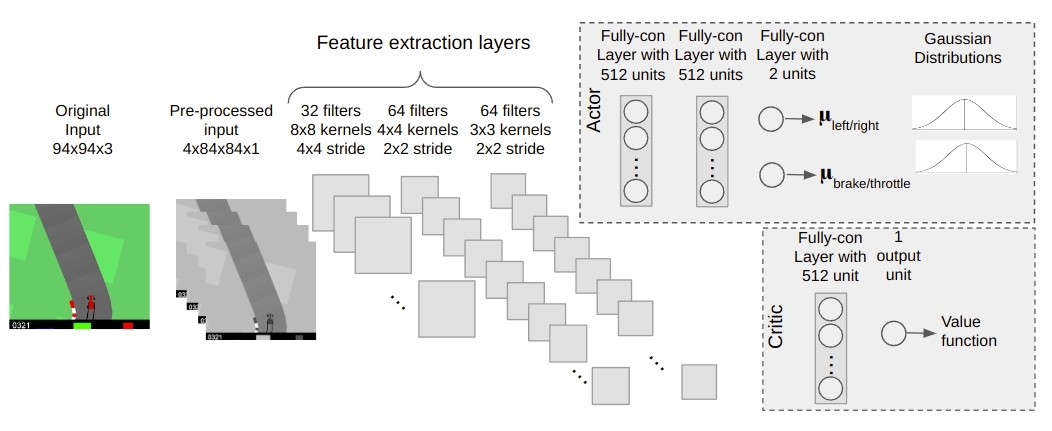}
\caption{Gaussian learner: the actor outputs the means of two diagonal Gaussian action distributions for steering and brake/throttle.}
\end{subfigure}

\vspace{0.8em}

\begin{subfigure}[t]{0.96\textwidth}
\centering
\includegraphics[width=\linewidth]{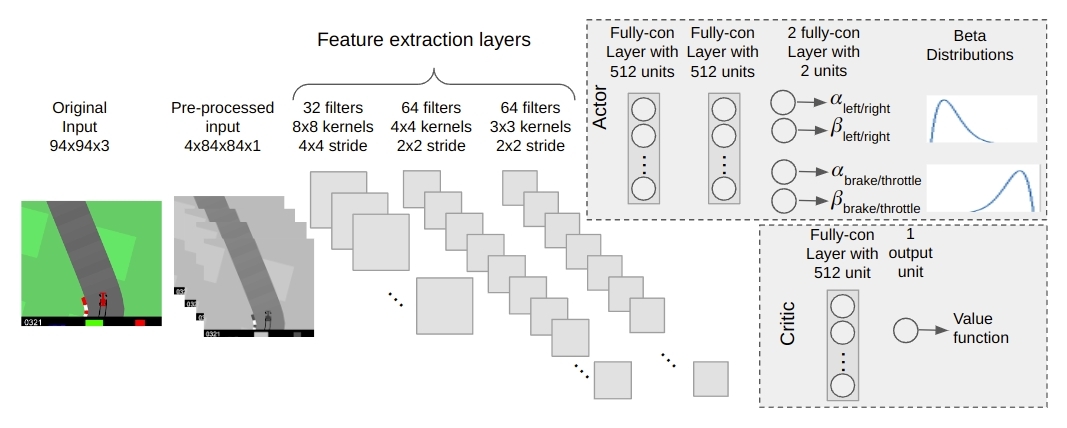}
\caption{Beta learner: the actor outputs $\alpha$ and $\beta$ for two independent bounded action distributions.}
\end{subfigure}
\caption{Visual actor--critic architectures used in this work. Both learners use the same image preprocessing and convolutional feature extractor. The actor head changes from Gaussian outputs in (a) to Beta-distribution parameters in (b), while the critic estimates a scalar value function. The disagreement ensemble remains Gaussian in all experiments.
}
\label{fig:architectures}
\end{figure}

For both Gaussian and Beta policies, the visual encoder comprises three convolutional layers followed by a 512-unit fully connected layer (Figure~\ref{fig:architectures}). The actor has two additional 512-unit fully connected layers, whereas the critic uses one separate 512-unit fully connected layer followed by a scalar output.

The Gaussian actor predicts the mean for steering and brake/throttle and uses a diagonal Gaussian distribution. Sampled actions outside $[-1,1]^2$ are clipped before execution. In BC, Eq.~\eqref{eq:bc} directly constrains the mean, while the distribution scale is not identified by the supervised target.

The Beta actor predicts four positive outputs, $(\alpha_1,\beta_1,\alpha_2,\beta_2)$, using a softplus transformation and parameters greater than one. Each Beta sample $x\in[0,1]$ is mapped to $a=2x-1$. Its deterministic mean is
\begin{equation}
\mu_{\mathrm{Beta}}(s)=2\frac{\alpha(s)}{\alpha(s)+\beta(s)}-1.
\end{equation}
Optimizing the mean changes both shape parameters and therefore also changes the variance.

\section{Experimental design}\label{sec:experimental}

\subsection{Staged implementation checks}

Before the CarRacing study, the implementation was checked in the two regimes originally emphasized by DRIL. Breakout combines image observations and discrete actions. LunarLanderContinuous combines a low-dimensional state and continuous actions. These experiments are not treated as new algorithmic contributions; they verify that the implementation can reproduce the expected qualitative behavior before combining visual input and continuous actions. Their numerical summary and corresponding figures are provided in Appendix~\ref{app:validation}.

\subsection{CarRacing task and preprocessing}

CarRacing-v0 generates a new top-down racing track for each episode \cite{brockman2016}. The original observation is a $96\times96$ RGB image. Each frame is converted to grayscale, resized to $84\times84$, and stacked with the previous three frames, producing a $4\times84\times84$ input. The stack supplies short-term motion information without an explicit speed sensor.

The track is divided into tiles. The agent receives $1000/N_{\mathrm{tiles}}$ for each newly visited tile and loses 0.1 per frame. An episode ends after all tiles are visited, after 1000 frames, or when the vehicle moves too far from the track. An average score above 900 over 100 consecutive episodes is the conventional solving criterion for this environment. The 100-episode evaluations use procedurally generated tracks that are distinct from the recorded demonstration trajectories, so they probe closed-loop behavior on new track realizations. For each retained policy, the reported mean and standard deviation are calculated over these 100 evaluation episodes.

\subsection{Clipped-action and bounded-action experts}

Two expert sources produce the demonstration datasets:
\begin{enumerate}[leftmargin=*]
    \item a PPO policy with a Gaussian action distribution. Approximately 73\% of sampled expert actions fall outside the valid interval in the analyzed trajectory, so the executed and stored actions are clipped to $[-1,1]^2$;
    \item a PPO policy with a Beta action distribution. Its support is mapped to $[-1,1]^2$, so every sampled action is bounded by construction.
\end{enumerate}
The selected Gaussian expert was reported at $897\pm41$ and the selected Beta expert at $913\pm26$ over 100 episodes. The Beta PPO result and expert training were previously published \cite{petrazzini2021}; they are used here as an expert-data source. For each source, 20 successful trajectories with episode score above 900 were collected. Experiments use either the first trajectory or all 20 trajectories.

\begin{figure}[t]
\centering
\includegraphics[width=\textwidth]{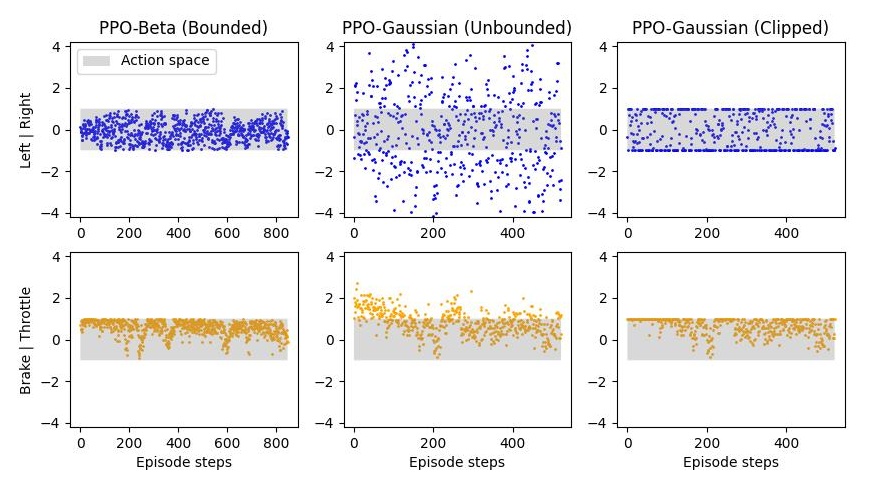}
\caption{Expert-action generation. The Beta expert produces bounded samples; the Gaussian expert produces many samples outside the action range; the clipped Gaussian actions are those executed in the environment and used in the clipped-action demonstration dataset.}
\label{fig:expert_actions}
\end{figure}

Figure~\ref{fig:expert_actions} illustrates the different action distributions. The clipped expert contains many saturated commands at the limits, whereas the bounded expert spreads actions within the valid interval.

\subsection{Factorial comparison}

The study crosses five experimental factors (Table~\ref{tab:factors}). The resulting policies are evaluated in both deterministic mode (distribution mean) and stochastic mode (sampling from the distribution). The same definitions are used for BC, DRIL-peak, and DRIL-final.

\begin{table}[t]
\caption{Experimental factors in the main CarRacing study. Every learner is evaluated in deterministic mode, using the distribution mean, and stochastic mode, sampling from the distribution. BC denotes the validation-selected behavior-cloning checkpoint; DRIL-peak is selected by the highest 10-episode training-score mean; DRIL-final is the last checkpoint.}\label{tab:factors}
\centering
\begin{tabularx}{\textwidth}{@{}lL@{}}
\toprule
Factor & Levels \\
\midrule
Expert demonstration actions & Clipped Gaussian; intrinsically bounded Beta \\
Number of expert trajectories & 1; 20 \\
Learner policy distribution & Gaussian; Beta \\
Retained training stage & BC; DRIL-peak; DRIL-final \\
Evaluation mode & Deterministic mean; stochastic sampling \\
\bottomrule
\end{tabularx}
\end{table}

The principal PPO and BC settings are listed in Table~\ref{tab:hyper}. Adam is used for optimization \cite{kingma2014}. The experimental record documents one main training run for each CarRacing configuration. Each retained policy is then evaluated over 100 consecutive procedurally generated tracks. Consequently, the reported mean and standard deviation quantify episode-to-episode variability conditional on a trained checkpoint; they do not quantify variability across independent retraining seeds.

\begin{table}[t]
\caption{Main CarRacing training settings. Ensemble members are Gaussian policies trained for 2000 epochs on bootstrap samples. The learner is first trained by BC and then updated by interleaved PPO and BC steps during DRIL.}\label{tab:hyper}
\centering
\begin{tabularx}{\textwidth}{@{}lYlY@{}}
\toprule
Parameter & Value & Parameter & Value \\
\midrule
BC learning rate & $2.5\times10^{-4}$ & BC mini-batch & 32 \\
BC data split & 80/20 & BC patience & 20 epochs \\
Ensemble size & 5 & Ensemble training & 2000 epochs \\
Uncertainty quantile & 0.98 & PPO learning rate & $3\times10^{-4}$, annealed \\
PPO rollout & 2048 steps & PPO epochs & 10 \\
PPO mini-batches & 32 & Discount $\gamma$ & 0.99 \\
GAE $\lambda$ & 0.95 & Entropy coefficient & 0 \\
Value-loss coefficient & 0.5 & Max. gradient norm & 0.5 \\
DRIL horizon & $\approx3\times10^6$ steps & Peak window & 10 episodes \\
\bottomrule
\end{tabularx}
\end{table}

The experimental comparison focuses on BC and DRIL because the objective is to isolate the effect of disagreement regularization under controlled changes in policy representation, demonstration source, and dataset size. Several recent imitation-learning methods discussed in Section 2, including energy-based, transformer, and diffusion policies, target multimodal or temporally extended action generation and require substantially different policy architectures and training protocols. Interactive approaches such as DAgger additionally require repeated expert labeling, while adversarial and hybrid RL--IL methods introduce different supervision or reward assumptions. Direct numerical comparison would therefore confound the effect of disagreement regularization with changes in the learning paradigm. Instead, BC provides the common supervised initialization and primary baseline for every experimental configuration, while the complete factorial evaluation isolates when DRIL improves upon that baseline.

The implementation is additionally evaluated in Breakout and LunarLanderContinuous to verify its qualitative behavior in the two observation/action regimes originally considered by DRIL before evaluating their combination in CarRacing.

\section{Results}\label{sec:results}

Across Figs.~\ref{fig:gclip-training}--\ref{fig:bbound-evaluation}, points represent individual episode scores. In the training plots, the orange curve is the 10-episode moving average used for DRIL-peak selection; in the evaluation plots, the orange horizontal line indicates the 100-episode mean.

\subsection{Gaussian learner policies}

Table~\ref{tab:gaussian} reports all Gaussian learner results. Under clipped-action demonstrations, stochastic BC has a low mean with one trajectory (30$\pm$67), but improves to 437$\pm$115 with 20 trajectories. Stochastic DRIL-peak increases the mean to 322$\pm$208 and 802$\pm$197, respectively. The final checkpoints fall to 39$\pm$79 and 218$\pm$108. The same pattern appears under bounded-action demonstrations: stochastic DRIL-peak reaches 341$\pm$246 and 720$\pm$289, whereas DRIL-final reaches only 184$\pm$130 and 240$\pm$129.

\begin{table}[t]
\caption{Gaussian learner evaluation on CarRacing. Values are mean $\pm$ standard deviation over 100 evaluation episodes. Bold values mark the highest mean scores within each evaluation mode and dataset regime.
}\label{tab:gaussian}
\centering
\small
\begin{tabular}{@{}llcccc@{}}
\toprule
Expert & Method & \multicolumn{2}{c}{1 trajectory} & \multicolumn{2}{c}{20 trajectories}\\
\cmidrule(lr){3-4}\cmidrule(lr){5-6}
& & Deterministic & Stochastic & Deterministic & Stochastic\\
\midrule
\multirow{3}{*}{Clipped}
& BC & \pmv{125}{113} & \pmv{30}{67} & \pmv{171}{124} & \pmv{437}{115}\\
& DRIL-peak & \pmv{166}{131} & \best{\pmv{322}{208}} & \pmv{423}{211} & \best{\pmv{802}{197}}\\
& DRIL-final & \pmv{41}{80} & \pmv{39}{79} & \pmv{229}{107} & \pmv{218}{108}\\
\midrule
\multirow{3}{*}{Bounded}
& BC & \pmv{194}{113} & \pmv{75}{47} & \pmv{617}{260} & \pmv{137}{70}\\
& DRIL-peak & \pmv{235}{123} & \best{\pmv{341}{246}} & \best{\pmv{656}{266}} & \best{\pmv{720}{289}}\\
& DRIL-final & \pmv{202}{128} & \pmv{184}{130} & \pmv{246}{120} & \pmv{240}{129}\\
\bottomrule
\end{tabular}
\end{table}

\begin{figure}[p]
\centering
\begin{subfigure}[t]{0.98\textwidth}
  \centering
  \includegraphics[width=\linewidth]{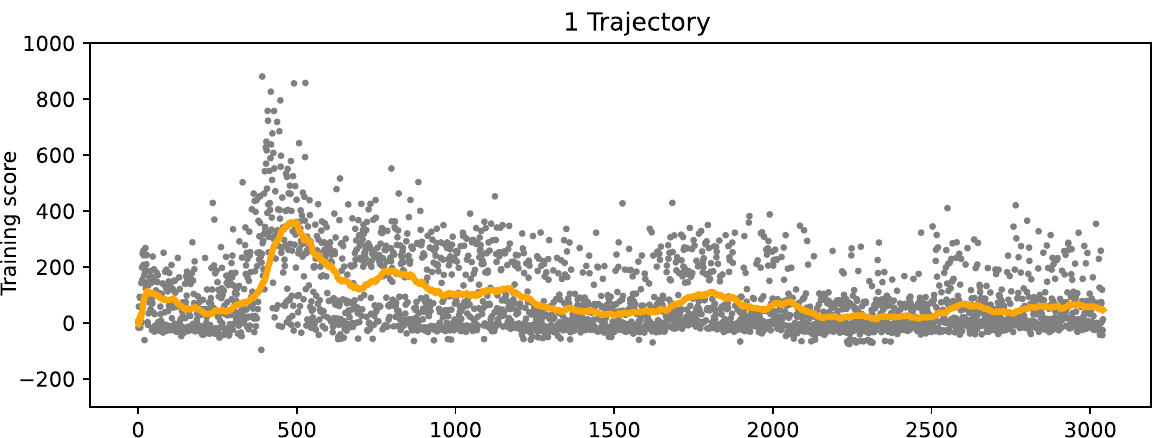}
  \caption{DRIL training with one demonstration.}
  \label{fig:gclip-train1}
\end{subfigure}

\vspace{0.7em}
\begin{subfigure}[t]{0.98\textwidth}
  \centering
  \includegraphics[width=\linewidth]{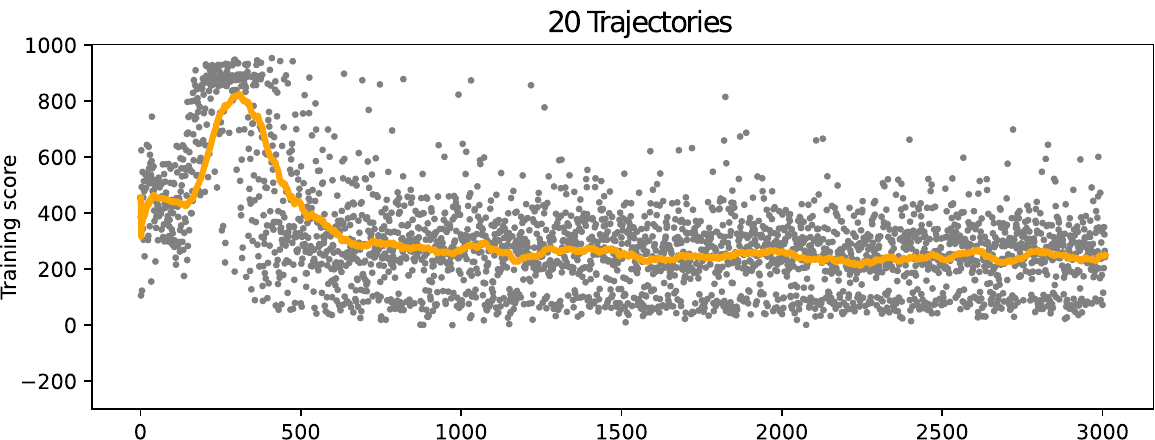}
  \caption{DRIL training with 20 demonstrations.}
  \label{fig:gclip-train20}
\end{subfigure}
\caption{Training dynamics of the Gaussian learner using clipped-action expert demonstrations. Panels (a) and (b) correspond to one and 20 demonstrations, respectively.}
\label{fig:gclip-training}
\end{figure}

\begin{figure}[p]
\centering
\begin{subfigure}[t]{0.98\textwidth}
  \centering
  \includegraphics[width=\linewidth]{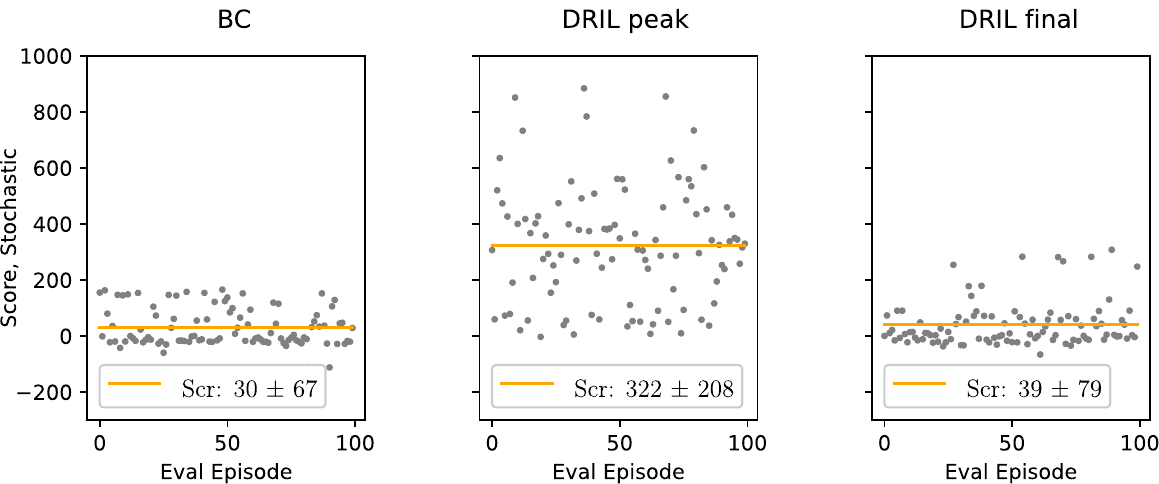}
  \caption{Stochastic evaluation with one demonstration.}
  \label{fig:gclip-eval1}
\end{subfigure}

\vspace{0.8em}
\begin{subfigure}[t]{0.98\textwidth}
  \centering
  \includegraphics[width=\linewidth]{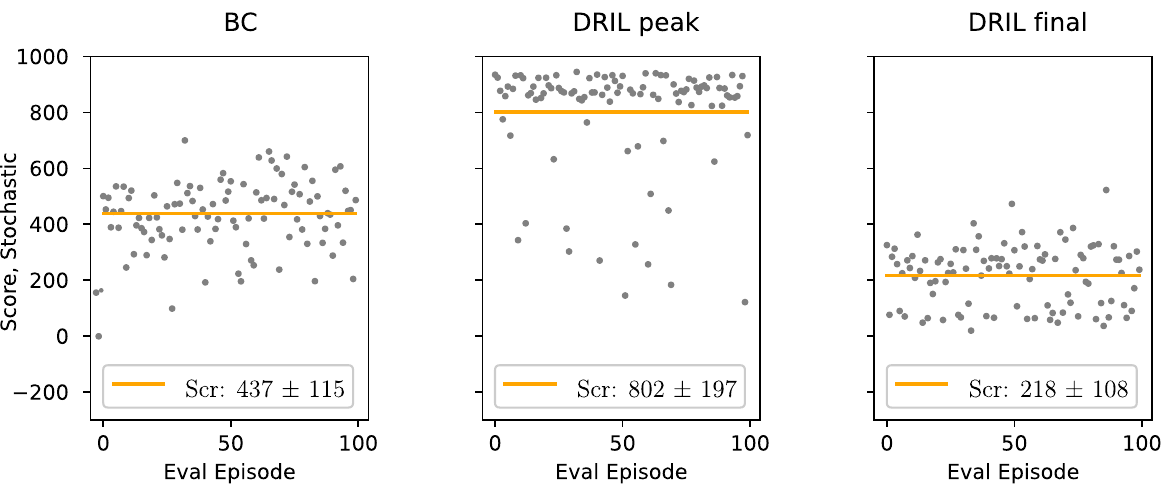}
  \caption{Stochastic evaluation with 20 demonstrations.}
  \label{fig:gclip-eval20}
\end{subfigure}
\caption{Stochastic 100-episode evaluation of the Gaussian learner using clipped-action expert demonstrations. Panels (a) and (b) correspond to one and 20 demonstrations, respectively, and compare BC, DRIL-peak, and DRIL-final. Deterministic results are reported in Table~\ref{tab:gaussian}.}
\label{fig:gclip-evaluation}
\end{figure}

\begin{figure}[p]
\centering
\begin{subfigure}[t]{0.98\textwidth}
  \centering
  \includegraphics[width=\linewidth]{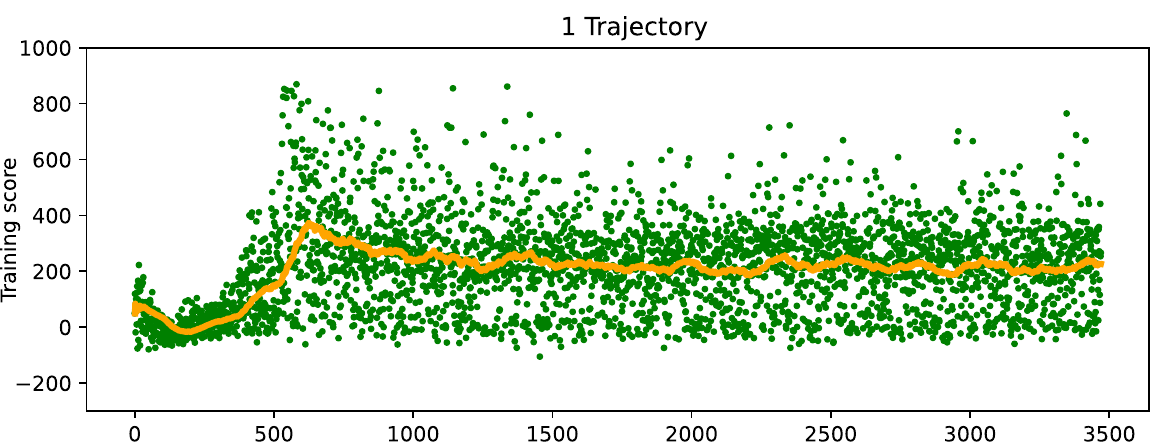}
  \caption{DRIL training with one demonstration.}
  \label{fig:gbound-train1}
\end{subfigure}

\vspace{0.7em}
\begin{subfigure}[t]{0.98\textwidth}
  \centering
  \includegraphics[width=\linewidth]{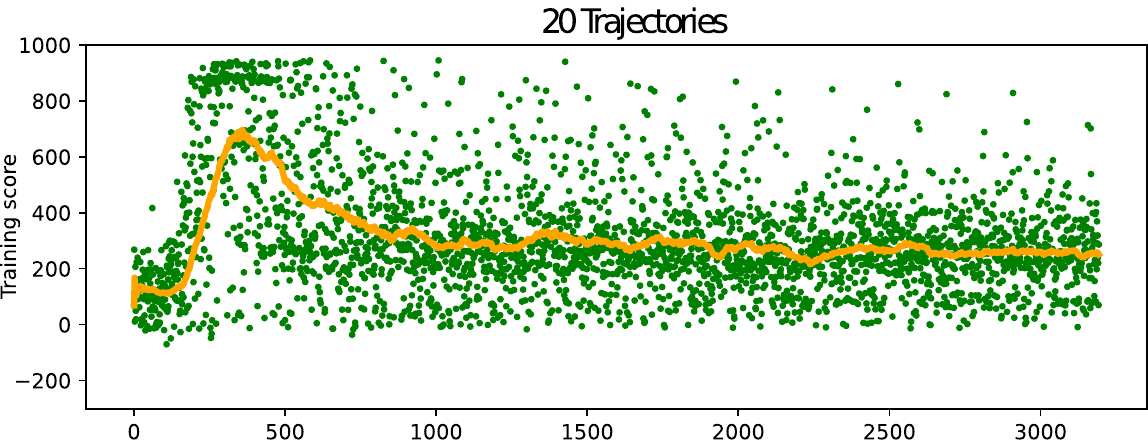}
  \caption{DRIL training with 20 demonstrations.}
  \label{fig:gbound-train20}
\end{subfigure}
\caption{Training dynamics of the Gaussian learner using bounded-action expert demonstrations. Panels (a) and (b) correspond to one and 20 demonstrations, respectively.}
\label{fig:gbound-training}
\end{figure}

\begin{figure}[p]
\centering
\begin{subfigure}[t]{0.98\textwidth}
  \centering
  \includegraphics[width=\linewidth]{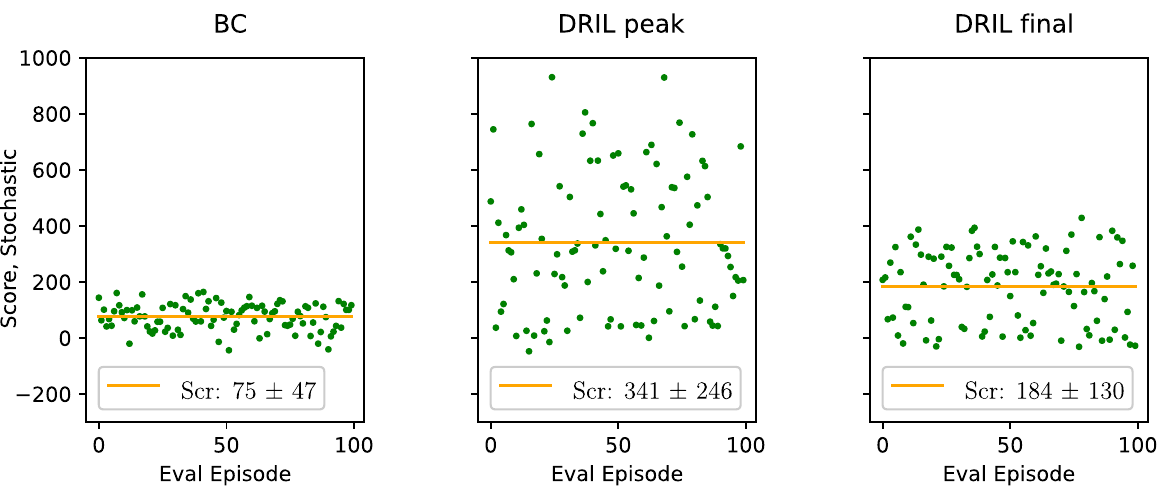}
  \caption{Stochastic evaluation with one demonstration.}
  \label{fig:gbound-eval1}
\end{subfigure}

\vspace{0.8em}
\begin{subfigure}[t]{0.98\textwidth}
  \centering
  \includegraphics[width=\linewidth]{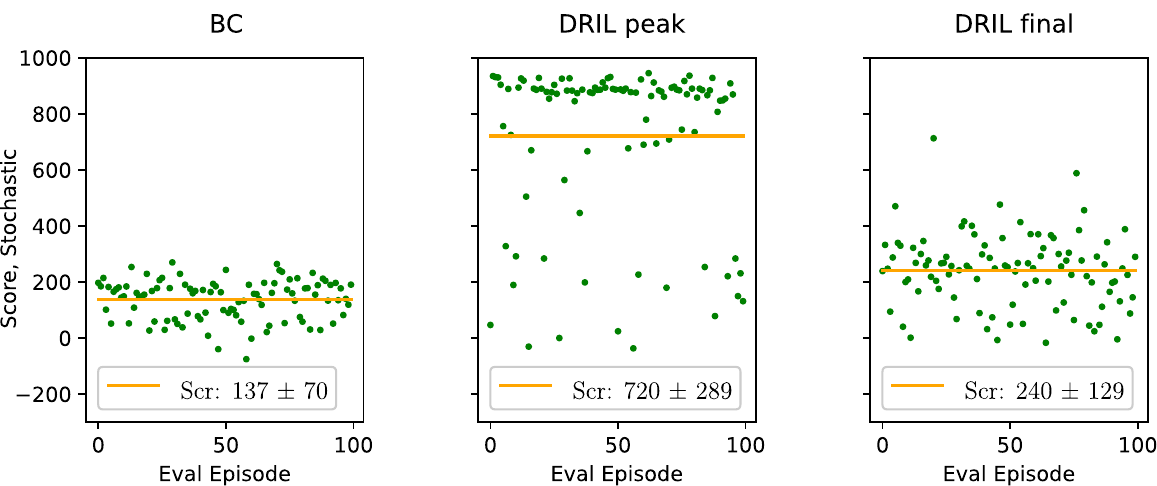}
  \caption{Stochastic evaluation with 20 demonstrations.}
  \label{fig:gbound-eval20}
\end{subfigure}
\caption{Stochastic 100-episode evaluation of the Gaussian learner using bounded-action expert demonstrations. Panels (a) and (b) correspond to one and 20 demonstrations, respectively, and compare BC, DRIL-peak, and DRIL-final. Deterministic results are reported in Table~\ref{tab:gaussian}.}
\label{fig:gbound-evaluation}
\end{figure}

For clipped-action demonstrations, the DRIL training traces and stochastic evaluation distributions are shown in Figs.~\ref{fig:gclip-training} and \ref{fig:gclip-evaluation}, respectively. The corresponding bounded-action results are shown in Figs.~\ref{fig:gbound-training} and \ref{fig:gbound-evaluation}. In both demonstration regimes, the score rises early in DRIL training and later declines. The 10-episode moving-average checkpoint therefore retains a substantially different policy from the final checkpoint.

\clearpage

\clearpage

\clearpage

\clearpage

The stochastic Gaussian peak policy is generally stronger than its deterministic counterpart, indicating that action sampling can help during evaluation.

\subsection{Beta learner policies}

Table~\ref{tab:beta} reports the Beta learner results. BC is much more competitive in stochastic mode than the Gaussian BC policy. With 20 trajectories, stochastic Beta BC reaches 758$\pm$237 for clipped demonstrations and 794$\pm$227 for bounded demonstrations. The latter is the highest mean in the 20-trajectory bounded-expert regime.

With one trajectory, stochastic DRIL-peak improves Beta BC from 216$\pm$108 to 348$\pm$230 for clipped demonstrations and from 161$\pm$156 to 200$\pm$184 for bounded demonstrations. With 20 trajectories, DRIL-peak remains below Beta BC: it reaches 572$\pm$280 and 738$\pm$277, respectively. All four DRIL-final stochastic means are between 99 and 267, showing a consistent decrease from the strongest intermediate checkpoints across both learner classes.

\begin{table}[t]
\caption{Beta learner evaluation on CarRacing with the fixed Gaussian disagreement ensemble. Values are mean $\pm$ standard deviation over 100 evaluation episodes.
Bold values mark the highest mean scores within each evaluation mode and dataset regime.
}
\label{tab:beta}
\centering
\small
\begin{tabular}{@{}llcccc@{}}
\toprule
Expert & Method & \multicolumn{2}{c}{1 trajectory} & \multicolumn{2}{c}{20 trajectories}\\
\cmidrule(lr){3-4}\cmidrule(lr){5-6}
& & Deterministic & Stochastic & Deterministic & Stochastic\\
\midrule
\multirow{3}{*}{Clipped}
& BC & \pmv{163}{67} & \pmv{216}{108} & \pmv{252}{143} & \best{\pmv{758}{237}}\\
& DRIL-peak & \best{\pmv{229}{142}} & \best{\pmv{348}{230}} & \pmv{210}{115} & \pmv{572}{280}\\
& DRIL-final & \pmv{122}{116} & \pmv{143}{111} & \best{\pmv{265}{165}} & \pmv{241}{140}\\
\midrule
\multirow{3}{*}{Bounded}
& BC & \pmv{147}{105} & \pmv{161}{156} & \best{\pmv{567}{239}} & \best{\pmv{794}{227}}\\
& DRIL-peak & \best{\pmv{195}{88}} & \best{\pmv{200}{184}} & \pmv{398}{225} & \pmv{738}{277}\\
& DRIL-final & \pmv{123}{125} & \pmv{99}{121} & \pmv{275}{144} & \pmv{267}{153}\\
\bottomrule
\end{tabular}
\end{table}

\begin{figure}[p]
\centering
\begin{subfigure}[t]{0.98\textwidth}
  \centering
  \includegraphics[width=\linewidth]{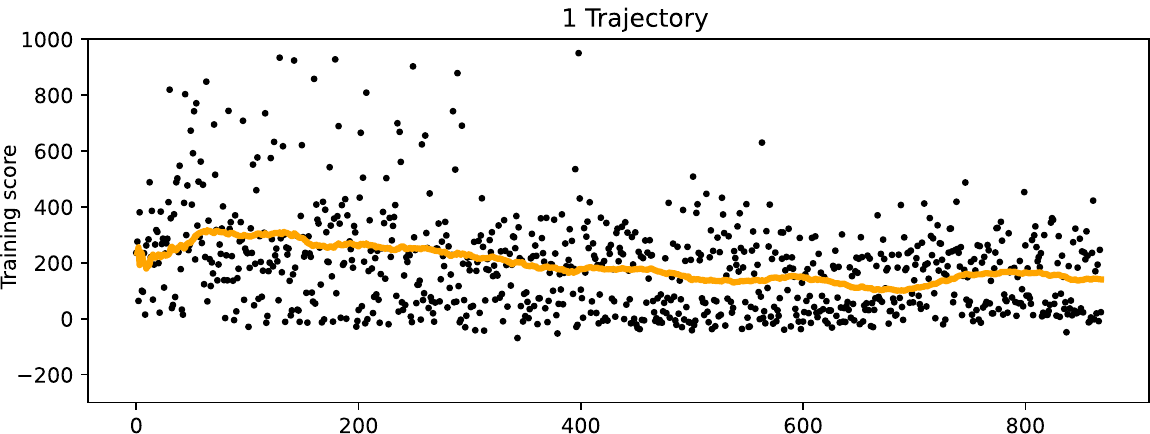}
  \caption{DRIL training with one demonstration.}
  \label{fig:bclip-train1}
\end{subfigure}

\vspace{0.7em}
\begin{subfigure}[t]{0.98\textwidth}
  \centering
  \includegraphics[width=\linewidth]{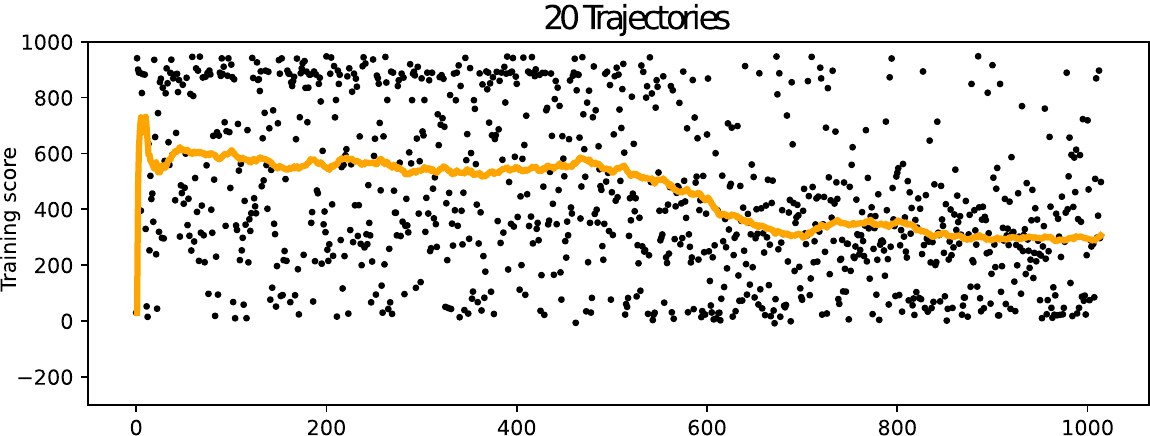}
  \caption{DRIL training with 20 demonstrations.}
  \label{fig:bclip-train20}
\end{subfigure}
\caption{Training dynamics of the Beta learner using clipped-action expert demonstrations and the fixed Gaussian disagreement ensemble. Panels (a) and (b) correspond to one and 20 demonstrations, respectively.}
\label{fig:bclip-training}
\end{figure}

\begin{figure}[p]
\centering
\begin{subfigure}[t]{0.98\textwidth}
  \centering
  \includegraphics[width=\linewidth]{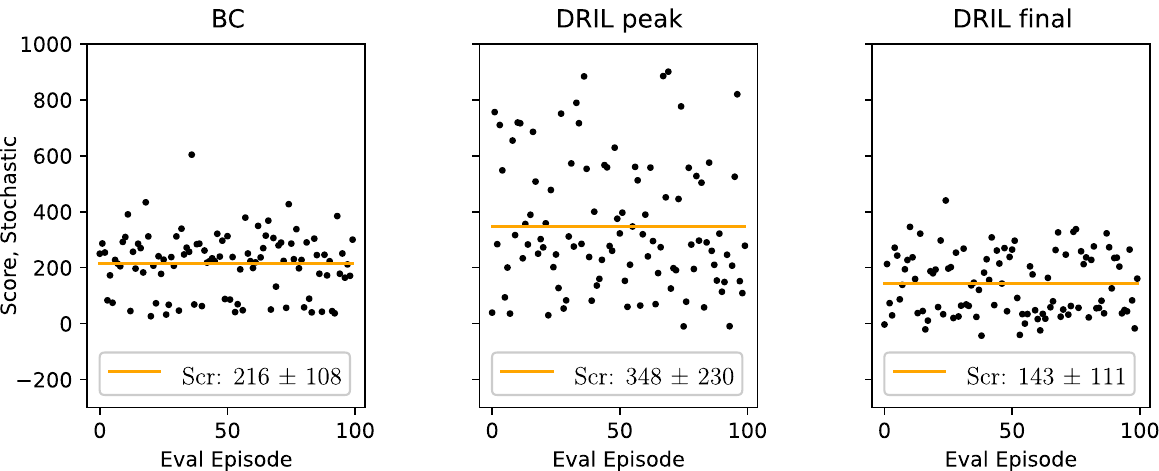}
  \caption{Stochastic evaluation with one demonstration.}
  \label{fig:bclip-eval1}
\end{subfigure}

\vspace{0.8em}
\begin{subfigure}[t]{0.98\textwidth}
  \centering
  \includegraphics[width=\linewidth]{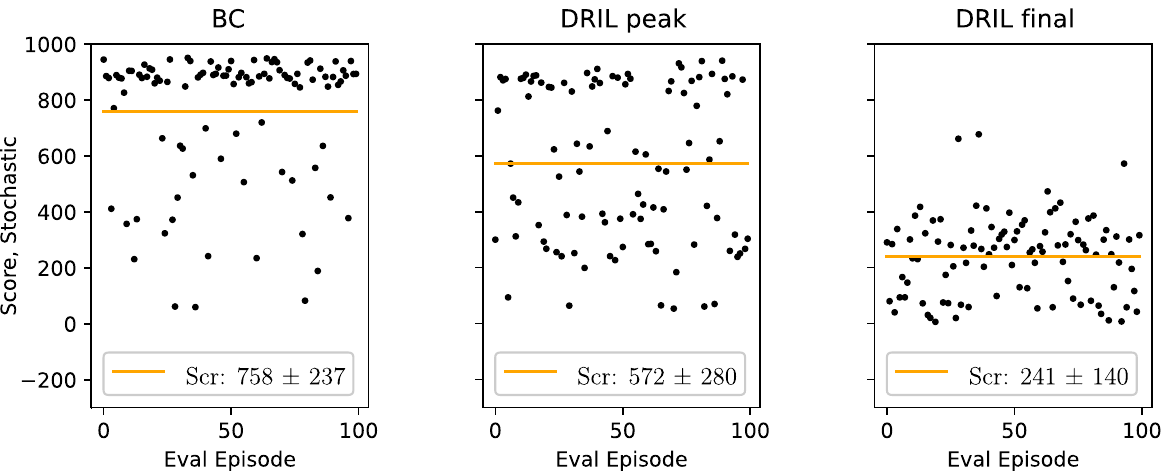}
  \caption{Stochastic evaluation with 20 demonstrations.}
  \label{fig:bclip-eval20}
\end{subfigure}
\caption{Stochastic 100-episode evaluation of the Beta learner using clipped-action expert demonstrations and the fixed Gaussian disagreement ensemble. Panels (a) and (b) correspond to one and 20 demonstrations, respectively, and compare BC, DRIL-peak, and DRIL-final. Deterministic results are reported in Table~\ref{tab:beta}.}
\label{fig:bclip-evaluation}
\end{figure}

\begin{figure}[p]
\centering
\begin{subfigure}[t]{0.98\textwidth}
  \centering
  \includegraphics[width=\linewidth]{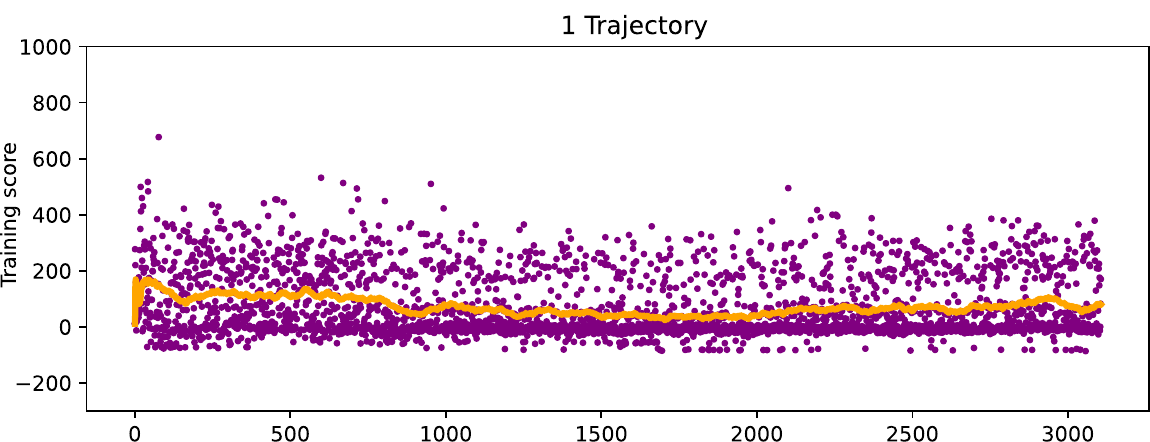}
  \caption{DRIL training with one demonstration.}
  \label{fig:bbound-train1}
\end{subfigure}

\vspace{0.7em}
\begin{subfigure}[t]{0.98\textwidth}
  \centering
  \includegraphics[width=\linewidth]{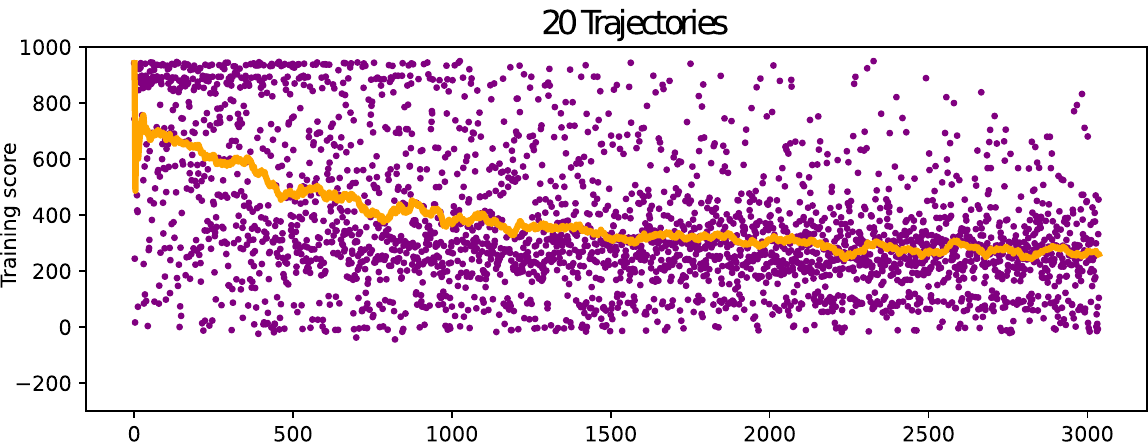}
  \caption{DRIL training with 20 demonstrations.}
  \label{fig:bbound-train20}
\end{subfigure}
\caption{Training dynamics of the Beta learner using bounded-action expert demonstrations and the fixed Gaussian disagreement ensemble. Panels (a) and (b) correspond to one and 20 demonstrations, respectively.}
\label{fig:bbound-training}
\end{figure}

\begin{figure}[p]
\centering
\begin{subfigure}[t]{0.98\textwidth}
  \centering
  \includegraphics[width=\linewidth]{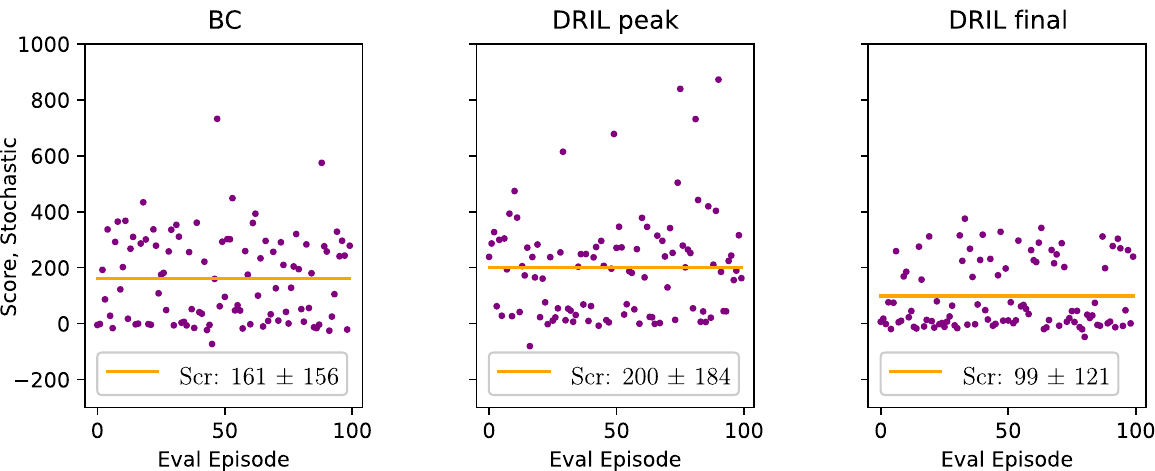}
  \caption{Stochastic evaluation with one demonstration.}
  \label{fig:bbound-eval1}
\end{subfigure}

\vspace{0.8em}
\begin{subfigure}[t]{0.98\textwidth}
  \centering
  \includegraphics[width=\linewidth]{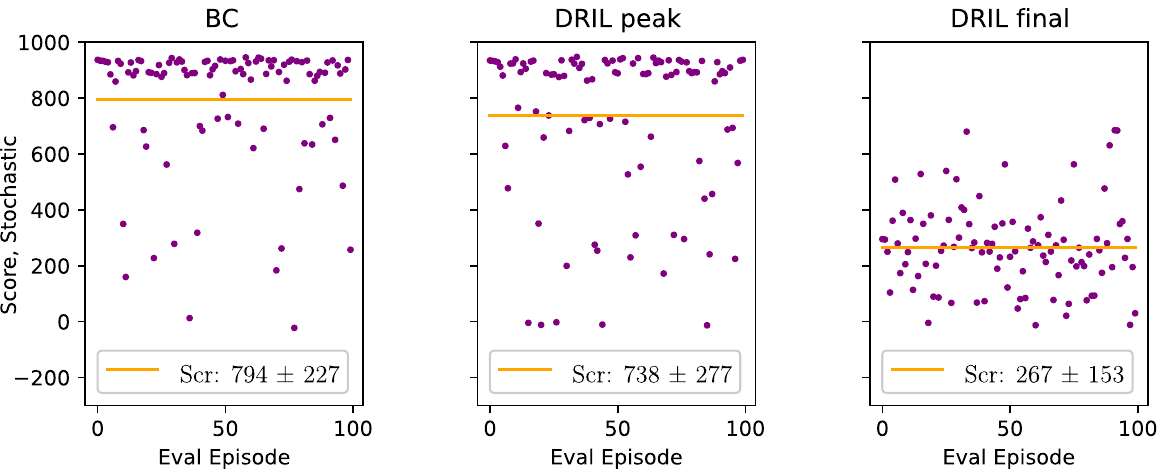}
  \caption{Stochastic evaluation with 20 demonstrations.}
  \label{fig:bbound-eval20}
\end{subfigure}
\caption{Stochastic 100-episode evaluation of the Beta learner using bounded-action expert demonstrations and the fixed Gaussian disagreement ensemble. Panels (a) and (b) correspond to one and 20 demonstrations, respectively, and compare BC, DRIL-peak, and DRIL-final. Deterministic results are reported in Table~\ref{tab:beta}.}
\label{fig:bbound-evaluation}
\end{figure}

For the Beta learner, clipped-action training and stochastic evaluation results are shown in Figs.~\ref{fig:bclip-training} and \ref{fig:bclip-evaluation}; bounded-action results are shown in Figs.~\ref{fig:bbound-training} and \ref{fig:bbound-evaluation}. The Beta BC policy often starts DRIL from a high score, especially with 20 demonstrations. During the subsequent PPO phase, performance decreases from the BC initialization. In the 20-trajectory bounded case, the highest retained DRIL checkpoint remains below BC, making BC the strongest method for that regime.

\clearpage

\clearpage

\clearpage

\clearpage

\subsection{Cross-factor comparison}

Table~\ref{tab:best} compares the strongest stochastic BC and DRIL-peak policy in each demonstration regime. DRIL-peak has the highest mean in three regimes. Relative to the strongest BC mean, the increases are 61.1\% for one clipped trajectory, 5.8\% for 20 clipped trajectories, and 111.8\% for one bounded trajectory. With 20 bounded trajectories, the best BC mean is 7.1\% higher than the best DRIL-peak mean. These percentages summarize the relative differences between the reported means.

\begin{table}[t]
\caption{Best stochastic BC and score-selected DRIL checkpoint in each demonstration regime, allowing the learner distribution to differ. Percent change is computed from the reported means.}\label{tab:best}
\centering
\small
\begin{tabularx}{\textwidth}{@{}llLLY@{}}
\toprule
Expert & Trajectories & Best BC & Best DRIL-peak & Change of DRIL peak \\
\midrule
Clipped & 1 & Beta: \pmv{216}{108} & Beta: \pmv{348}{230} & +61.1\% \\
Clipped & 20 & Beta: \pmv{758}{237} & Gaussian: \pmv{802}{197} & +5.8\% \\
Bounded & 1 & Beta: \pmv{161}{156} & Gaussian: \pmv{341}{246} & +111.8\% \\
Bounded & 20 & Beta: \pmv{794}{227} & Beta: \pmv{738}{277} & $-7.1$\% \\
\bottomrule
\end{tabularx}
\end{table}

Three interactions stand out. First, stochastic evaluation is usually better than deterministic evaluation for the strongest checkpoints, particularly for Beta BC and Gaussian DRIL-peak.
Second, the preferred learner distribution depends on the demonstration regime rather than solely on the expert policy family. For example, with one bounded-action trajectory generated by the Beta expert, the Gaussian learner achieves the strongest DRIL-peak result, whereas with 20 bounded-action trajectories the Beta learner provides the strongest overall result.
Third, more demonstrations reduce the headroom for DRIL when BC is already strong. The 20-trajectory bounded dataset produces a Beta BC policy with a high mean, and its mean remains above the retained DRIL checkpoints.

\subsection{Reward response with the Beta learner}

\begin{figure}[p]
\centering
\includegraphics[width=0.98\textwidth]{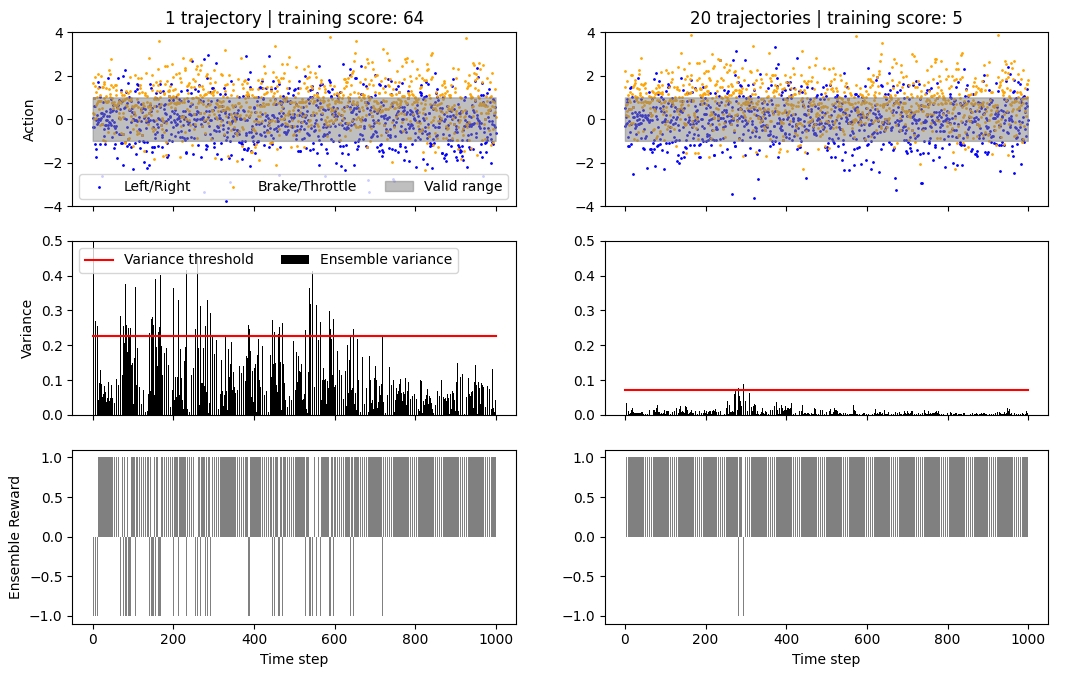}
\caption{Disagreement signal during the first 1000 interaction steps of the Gaussian learner. The left column uses one expert trajectory and the right column uses 20 trajectories. The top rows show steering and brake/throttle samples with the valid action interval shaded; the middle rows show scalar ensemble uncertainty and the 98th-percentile threshold; the bottom rows show the resulting binary reward. Sampled Gaussian actions frequently extend beyond the valid interval before clipping, and occasional uncertainty values cross the threshold.}
\label{fig:signal_gaussian}
\end{figure}

\begin{figure}[p]
\centering
\includegraphics[width=0.98\textwidth]{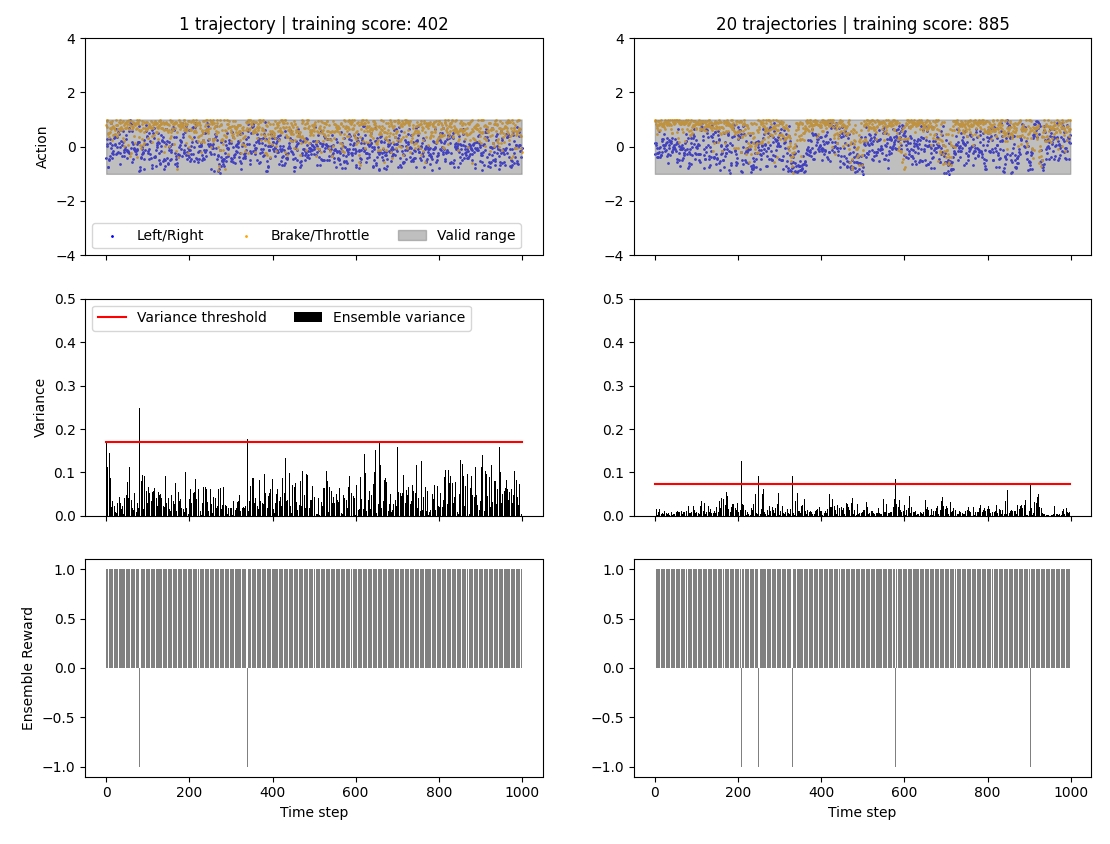}
\caption{Disagreement signal during the first 1000 interaction steps of the Beta learner using the same fixed Gaussian ensemble. The left column uses one expert trajectory and the right column uses 20 trajectories. The top rows show bounded steering and brake/throttle actions; the middle rows show scalar ensemble uncertainty and the 98th-percentile threshold; the bottom rows show the binary reward. Threshold crossings are infrequent, so the reward remains positive over most of the rollout.
}
\label{fig:signal_beta}
\end{figure}

Figures~\ref{fig:signal_gaussian} and \ref{fig:signal_beta} contrast the first 1000 DRIL steps for Gaussian and Beta learners. For the Gaussian learner, many sampled actions lie outside the valid range before clipping, and occasional ensemble variances exceed the threshold. For the Beta learner, all actions are bounded. More importantly, the ensemble variance often remains below the threshold even when the episode score indicates low task performance or departure from the track. The resulting reward remains close to $+1$ for most steps.

\FloatBarrier

A nearly constant reward gives PPO limited information for distinguishing actions during policy improvement, while PPO updates continue to move the actor away from the BC initialization. This explains the observed decrease in Beta-policy performance during DRIL training. Bounded actions remain beneficial for BC, but they also change the interaction between the learner and the fixed uncertainty estimator.

\section{Discussion}\label{sec:discussion}

\subsection{Checkpoint selection and training dynamics}

The difference between DRIL-peak and DRIL-final shows that checkpoint selection is an important part of the experimental analysis. For clipped demonstrations with 20 trajectories, the stochastic Gaussian mean falls from 802 at the selected peak to 218 at the final checkpoint. For bounded demonstrations with 20 trajectories, it falls from 720 to 240. Similar declines occur for Beta learners. Reporting both checkpoints captures the transient improvement and the subsequent evolution of the policy, providing a more complete characterization of DRIL training.

This evolution may reflect retention dynamics between the BC and PPO objectives. Candidate mechanisms include the binary reward, the balance and ordering of BC and PPO updates, value-function learning, the fixed uncertainty threshold, and changes in the learner distribution.
Additional experiments examined modifications to the disagreement reward, larger ensembles, and wider ensemble networks, but these changes did not eliminate the observed degradation.
The score-selected checkpoint nevertheless provides a consistent way to identify the strongest intermediate policy in simulation.

\subsection{Policy support and ensemble response}

The Beta policy has two clear advantages for BC. Every sampled action lies within the actuator range, and optimizing $\alpha$ and $\beta$ changes both the policy mean and concentration. This likely explains why stochastic Beta BC is strong with 20 demonstrations. The Gaussian BC objective constrains the mean but leaves the supervised objective with limited control over stochastic spread, which contributes to the lower one-trajectory stochastic mean.

The same boundedness changes DRIL's operating conditions. DRIL assumes that ensemble disagreement separates familiar from unfamiliar state--action pairs. In the Beta experiments, bounded rollouts can obtain low task scores without creating large action-direction variance in the fixed Gaussian ensemble.
This shows that keeping actions within the valid actuator range does not necessarily make the disagreement signal informative about whether the learner has moved outside the demonstrated distribution.
A policy can remain within actuator limits and still visit states that require a negative uncertainty reward.

\subsection{Expert demonstrations and dataset size}

The clipped and bounded experts differ both in action support and in the policy that generated the trajectories. Saturated actions in the clipped dataset can simplify some supervised mappings because many targets lie at $-1$ or $+1$, but they can also create abrupt control commands. The bounded expert produces smoother interior actions and higher expert evaluation performance. The two demonstration sources favor different learner/checkpoint combinations: the largest score, 802$\pm$197, comes from a Gaussian DRIL-peak learner trained on 20 clipped trajectories, while the strongest pure BC result, 794$\pm$227, comes from a Beta learner trained on 20 bounded trajectories.

Increasing the dataset from one to 20 trajectories generally improves BC, as expected. DRIL provides its clearest relative benefit when one trajectory leaves large coverage gaps. With 20 bounded demonstrations, BC is already strong and DRIL has little room to improve. This pattern is consistent with the motivation for disagreement regularization: it is most useful when demonstration coverage is sparse and the initial cloned policy benefits from an additional online signal.

\subsection{Relation to current imitation-learning methods}

Modern generative policies can represent multimodal or temporally extended actions more effectively than the explicit Gaussian and Beta actors studied here \cite{florence2022,shafiullah2022,chi2023,lee2024}. Covariate shift can still occur because a highly expressive policy may encounter out-of-distribution states. Recent work therefore combines expressive policies with active data collection or synthetic augmentation \cite{zhang2024dmd}. The present findings remain relevant because any uncertainty-driven correction mechanism must be calibrated to the chosen policy representation.
A richer multimodal policy can assign high probability to several distinct but valid actions for the same observation. In that case, disagreement between policy outputs may reflect legitimate behavioral ambiguity rather than epistemic uncertainty, making action-level disagreement harder to interpret as an out-of-distribution signal.

\section{Conclusion}\label{sec:conclusion}

This paper examined disagreement-regularized imitation learning in a regime that combines stacked visual observations with continuous bounded actions. The study crossed learner distribution, expert-action source, demonstration count, evaluation mode, and checkpoint choice. Score-selected DRIL checkpoints obtained the highest stochastic mean in three of four demonstration regimes, with the largest relative gains occurring in one-trajectory settings. When 20 bounded-action demonstrations were available, stochastic Beta BC was the strongest method, showing that an intrinsically bounded policy can provide a competitive alternative to online disagreement regularization when demonstration coverage is broader.

The results also clarify the interaction between policy support and the disagreement reward. Gaussian and Beta learners induce different rollout distributions, and the fixed Gaussian ensemble responds differently to them. For Beta learners, the reward often remains positive over long portions of the rollout, while the strongest intermediate checkpoints occur early in training. Reporting BC, score-selected, and final checkpoints therefore reveals complementary stages of the learning process and shows when DRIL adds value beyond the cloned initialization.

Future work will extend the complete factorial study across independent training seeds and a shared set of evaluation-track seeds, construct matched experts to separate action support from expert quality, and investigate policy-matched ensembles for Beta learners. Further directions include reward-free checkpoint criteria based on held-out demonstrations or learned success estimators, continuous or representation-level disagreement signals, explicit counterexample states, and comparisons with DAgger, energy-based, transformer, and diffusion policies. Evaluation in scenario-rich closed-loop benchmarks such as CARLA and Bench2Drive will further test how the observed interactions transfer to urban traffic, route commands, weather variation, and safety-critical events. Overall, the study shows that DRIL can be especially effective in few-demonstration visual continuous control when the ensemble reward remains informative throughout training.

\clearpage
\appendix
\section{Implementation-validation experiments}\label{app:validation}

Before the CarRacing adaptation, the implementation was checked in the two observation/action regimes evaluated by the original DRIL study: visual observations with discrete actions in Breakout and low-dimensional observations with continuous actions in LunarLanderContinuous. These experiments verify the qualitative behavior of the implementation before combining visual observations and continuous actions in the main study.

\begin{table}[t]
\caption{Implementation checks in the two regimes considered by the original DRIL study. Each entry is the mean score over 100 evaluation episodes. The expert threshold is the task score used to define successful or expert-level performance in the corresponding experiment.}
\label{tab:implementation_checks}
\centering
\begin{tabular}{@{}llccc@{}}
\toprule
Environment & Demonstrations & Expert threshold & BC & DRIL \\
\midrule
Breakout & 1 trajectory & 300 & 5 & 355 \\
Breakout & 3 trajectories & 300 & 7 & 339 \\
LunarLanderContinuous & 1 trajectory & 200 & 194 & 257 \\
LunarLanderContinuous & 20 trajectories & 200 & 232 & 228 \\
\bottomrule
\end{tabular}
\end{table}

\begin{figure}[p]
\centering
\begin{subfigure}[t]{0.88\textwidth}
\centering
\includegraphics[width=\linewidth]{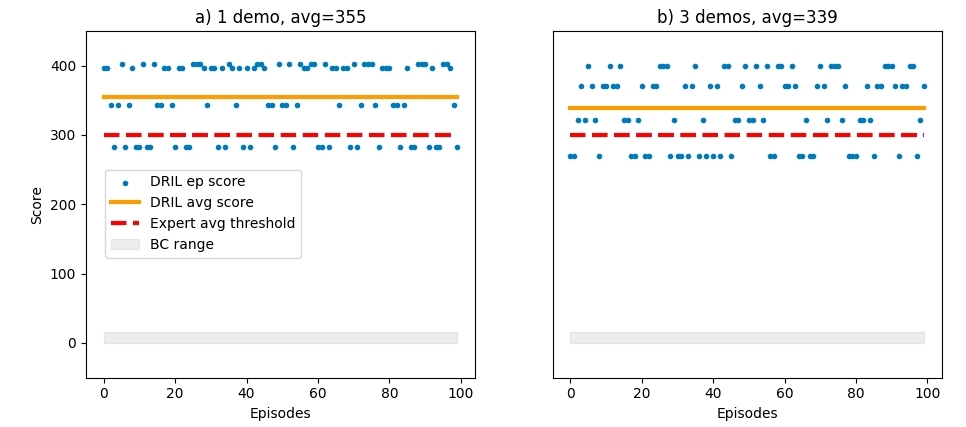}
\caption{Breakout evaluation. Each point is the score from one of 100 consecutive episodes. The left panel uses one expert trajectory and the right panel uses three trajectories. Horizontal lines show the DRIL mean, the BC mean, and the expert-level threshold of 300. DRIL reaches means of 355 and 339, compared with BC means of 5 and 7.}
\end{subfigure}
\vspace{0.8cm}

\begin{subfigure}[t]{0.88\textwidth}
\centering
\includegraphics[width=\linewidth]{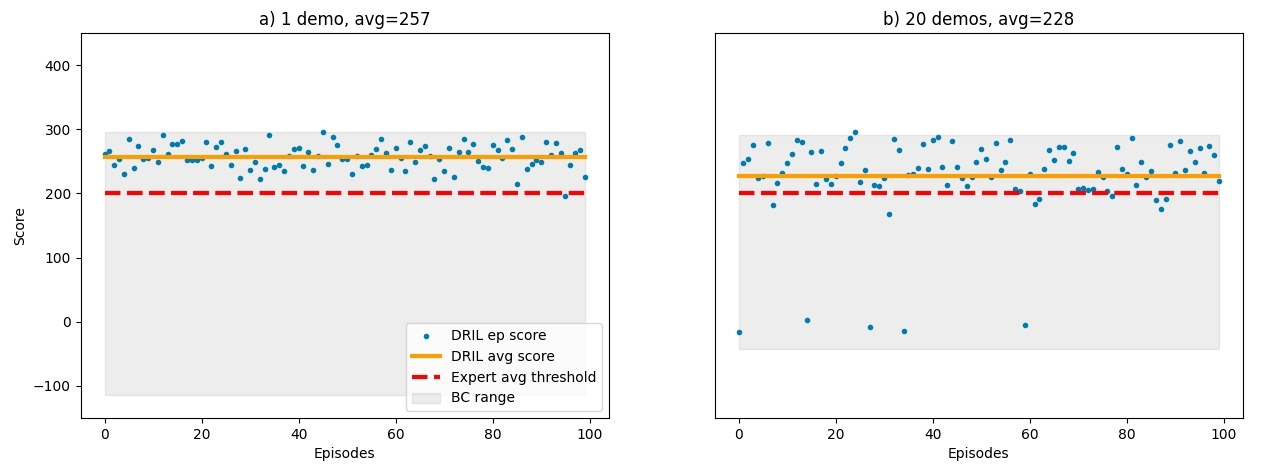}
\caption{LunarLanderContinuous evaluation. Each point is the score from one of 100 consecutive episodes. The left panel uses one expert trajectory and the right panel uses 20 trajectories. Horizontal lines show the DRIL mean, the BC mean, and the expert-level threshold of 200. DRIL changes the one-trajectory mean from 194 to 257 and the 20-trajectory mean from 232 to 228.}
\end{subfigure}
\caption{Implementation-validation results. The figures show the qualitative behavior of DRIL in the original discrete-visual and continuous-low-dimensional regimes before the image-based continuous-control study.}
\label{fig:implementation_validation}
\end{figure}
\FloatBarrier

\section*{Statements and Declarations}

\textbf{Funding:}
This work was partially funded by the National Council for Scientific and Technological Development - CNPq, Brazil (Grant No. 420148/2025-6) and in part by the Coordena\c{c}\~ao de Aperfei\c{c}oamento de Pessoal de N\'ivel Superior -- Brasil (CAPES) -- Finance Code 001.

\textbf{Competing interests:} The authors declare that they have no competing interests.

\textbf{Ethics approval:} Not applicable. The study uses simulated control environments and does not involve human participants or animals.

\textbf{Data and code availability:} Code for the DRIL adaptation is available at \url{https://github.com/igbp/dril}. Code associated with the previously published PPO-Beta expert is available at \url{https://github.com/igbp/SSCI}.

\textbf{Author contributions:} Irving Giovani Bronzatti Petrazzini contributed to methodology, software, investigation, validation, formal analysis, visualization, data curation, and writing--original draft. Eric Aislan Antonelo contributed to conceptualization, methodology, supervision, project administration, formal analysis, and writing--review and editing. Both authors read and approved the manuscript.

\bibliography{references}

\end{document}